\documentclass{article}

\usepackage{arxiv}

\usepackage[utf8]{inputenc} 
\usepackage[T1]{fontenc}    
\usepackage{hyperref}       
\usepackage{url}            
\usepackage{booktabs}       
\usepackage{amsfonts}       
\usepackage{nicefrac}       
\usepackage{microtype}      

\hypersetup{
    colorlinks=true,
    citecolor=black,
    linkcolor=blue,
    filecolor=magenta,      
    urlcolor=cyan,
    pdftitle={sbi - a toolkit for simulation-based inference},
    }

\title{\texttt{sbi} - a toolkit for simulation-based inference}

\author{
  Álvaro Tejero-Cantero\thanks{equally contributing.}$\; ^{\, ,\, 1},\;$
  Jan Boelts$^{*,\, 1},\;$ 
  Michael Deistler$^{*,\, 1},\;$ 
  Jan-Matthis Lueckmann$^{*,\, 1},\;$ \\ \And
  Conor Durkan$^{*,\, 2}$,
  Pedro J. Gonçalves$^{1,\, 3},\;$ 
  David S. Greenberg$^{1,\, 4},\;$ 
  Jakob H. Macke$^{1,\, 5,\, 6}$\\ [.3cm]
  Computational Neuroengineering, TU Munich$^1$, 
  School of Informatics, University of Edinburgh$^2$, \\
  Neural Systems Analysis, Center of Advanced European Studies and Research (caesar), Bonn$^3$, \\
  Model-Driven Machine Learning, Centre for Materials and Coastal Research, \\ 
  Helmholtz-Zentrum Geesthacht$^4$, \\
  Machine Learning in Science, University of Tübingen$^5$, \\
  Empirical Inference, Max Planck Institute for Intelligent Systems, Tübingen$^6$ \\[0.1cm]
  \{alvaro.tejero, jan.boelts, michael.deistler, jan-matthis.lueckmann\}@tum.de
}

\begin{document}
\maketitle

\begin{abstract}
Scientists and engineers employ stochastic numerical simulators to model empirically observed phenomena. In contrast to purely statistical models, simulators express scientific principles that provide powerful inductive biases, improve generalization to new data or scenarios and allow for fewer, more interpretable and domain-relevant parameters. 
Despite these advantages, tuning a simulator’s parameters so that its outputs match data is challenging. Simulation-based inference (SBI) seeks to identify parameter sets that a) are compatible with prior knowledge and b) match empirical observations. Importantly, SBI does not seek to recover a single ‘best’ data-compatible parameter set, but rather to identify all high probability regions of parameter space that explain observed data, and thereby to quantify parameter uncertainty. In Bayesian terminology, SBI aims to retrieve the posterior distribution over the parameters of interest. In contrast to conventional Bayesian inference, SBI is also applicable when one can run model simulations, but no formula or algorithm exists for evaluating the probability of data given parameters, i.e. the likelihood.

We present \texttt{sbi}, a PyTorch-based package\footnote{\texttt{sbi} is available at \href{http://mackelab.org/sbi}{\texttt{mackelab.org/sbi}} under the AGPLv3 license.} that implements SBI algorithms based on neural networks. \texttt{sbi} facilitates inference on black-box simulators for practising scientists and engineers by providing a unified interface to state-of-the-art algorithms together with documentation and tutorials.
\end{abstract}


\section{Motivation}
Bayesian inference is a principled approach for determining parameters consistent with empirical observations: Given a prior over parameters, a stochastic simulator, and observations, it returns a posterior distribution. In cases where the simulator likelihood \emph{can} be evaluated, many methods for approximate Bayesian inference exist (e.g., \cite{metropolis1953, neal2003, graham2017, le2016, baydin2020}). For more general simulators, however, evaluating the likelihood of data given parameters might be computationally intractable. Traditional algorithms for this `likelihood-free' setting \cite{cranmer2019} are based on Monte-Carlo rejection \cite{pritchard1999, sisson2007}, an approach known as  \emph{Approximate Bayesian Computation} (\textsc{ABC}). More recently, algorithms based on neural networks have been developed \cite{papamakarios2016, lueckmann2017, papamakarios2019a, greenberg2019, hermans2019}. These algorithms are not based on rejecting simulations, but rather train deep neural conditional density estimators or classifiers on simulated data. 
To aid in effective application of these algorithms to a wide range of problems, \texttt{sbi} closely integrates with PyTorch and offers state-of-the-art neural network-based SBI algorithms \cite{papamakarios2019a, hermans2019,greenberg2019} with flexible choice of network architectures and flow-based density estimators. With \texttt{sbi}, researchers can easily implement new neural inference algorithms, benefiting from the infrastructure to manage simulators and a unified posterior representation. Users, in turn, can profit from a single inference interface that allows them to either use their own custom neural network, or choose from a growing library of preconfigured options provided with the package.

\subsection{Related software and use in research}
We are aware of several mature packages that implement SBI algorithms. \texttt{elfi} \cite{elfi2018} is a package offering BOLFI, a Gaussian process-based algorithm \cite{gutmann2015}, and some classical ABC algorithms. The package \texttt{carl} \cite{louppe2016} implements the algorithm described in \cite{cranmer2015carl}. Two other SBI packages, currently under development, are \texttt{hypothesis} \cite{hypothesis-repo} and \texttt{pydelfi} \cite{pydelfi-repo}. \texttt{pyabc} \cite{Klinger2018} and \texttt{ABCpy} \cite{abcpy-repo} are two packages offering a diversity of ABC algorithms.

\texttt{sbi} is closely integrated with PyTorch \cite{paszke2019} and uses \texttt{nflows} \cite{nflows-repo} for flow-based density estimators. \texttt{sbi} builds on experience accumulated developing \texttt{delfi} \cite{delfi-repo}, which it succeeds. \texttt{delfi} was based on \texttt{theano} \cite{theano} (development discontinued) and developed both for SBI research \cite{greenberg2019, lueckmann2017} and for scientific applications \cite{goncalves2019}. The \texttt{sbi} codebase started as a fork of \texttt{lfi} \cite{lfi-repo}, developed for \cite{durkan2020}.

\section{Description}

\texttt{sbi} currently implements three families of neural inference algorithms:

\begin{itemize}
\item Sequential Neural \emph{Posterior} Estimation (SNPE) trains a deep neural density estimator that directly estimates the posterior distribution of parameters given data. Afterwards, it can sample parameter sets from the posterior, or evaluate the posterior density on any parameter set. Currently, SNPE-C \cite{greenberg2019} is implemented in \texttt{sbi}. 

\item Sequential Neural \emph{Likelihood} Estimation (SNLE) \cite{papamakarios2019a} trains a deep neural density estimator of the likelihood, which then allows to sample from the posterior using e.g. MCMC.

\item  Sequential Neural \emph{Ratio} Estimation (SNRE) \cite{hermans2019, durkan2020} trains a classifier to estimate density ratios, which in turn can be used to sample from the posterior e.g. with MCMC.
\end{itemize}
The inference step returns a \texttt{NeuralPosterior} object that represents the uncertainty about the parameters conditional on an observation, i.e. the posterior distribution. This object can be sampled from —and if the chosen algorithm allows, evaluated— with the same API as a standard PyTorch probability distribution.

An important challenge in making SBI algorithms usable by a broader community is to deal with diverse, often pre-existing, complex simulators. \texttt{sbi} works with any simulator as long as it can be wrapped in a Python callable. Furthermore, \texttt{sbi} ensures that custom simulators work well with neural networks, e.g. by performing automatic shape inference, standardizing inputs or handling failed simulations. To maximize simulator performance, \texttt{sbi} leverages vectorization where available and optionally parallelizes simulations using \texttt{joblib} \cite{joblib}. Moreover, if dimensionality reduction of the simulator output is desired, \texttt{sbi} can use a trainable summarizing network to extract relevant features from raw simulator output and spare the user manual feature engineering.

In addition to the full-featured interface, \texttt{sbi} provides also a \emph{simple} interface which consists of a single function call with reasonable defaults. This allows new users to get familiarized with simulation-based inference and quickly obtain results without having to define custom networks or tune hyperparameters.

With \texttt{sbi}, we aim to support scientific discovery and computational engineering by making Bayesian inference applicable to the widest class of models (simulators with no likelihood available), and practical for complex problems. We have designed an open architecture and adopted community-oriented development practices in order to invite other machine-learning researchers to join us in this long-term vision.

\subsection*{Acknowledgements}
This work has been supported by the German Federal Ministry of Education and Research (BMBF, project `ADIMEM', FKZ 01IS18052 A-D), the German Research Foundation (DFG) through  SFB 1089 `Synaptic Microcircuits', SPP 2041 `Computational Connectomics' and Germany’s Excellence Strategy – EXC-Number 2064/1 – Project number 390727645.

Conor Durkan was supported by the EPSRC Centre for Doctoral Training in Data Science, funded by the UK Engineering and Physical Sciences Research Council (grant EP/L016427/1) and the University of Edinburgh.

We are grateful to Artur Bekasov, George Papamakarios and Iain Murray for making \texttt{nflows} \cite{nflows-repo} available, a package for normalizing flow-based density estimation which \texttt{sbi} leverages extensively.

\bibliographystyle{unsrt}  

\bibliography{paper}

\begin{thebibliography}{10}

\bibitem{metropolis1953}
Nicholas Metropolis, Arianna~W Rosenbluth, Marshall~N Rosenbluth, Augusta~H
  Teller, and Edward Teller.
\newblock Equation of state calculations by fast computing machines.
\newblock {\em The journal of chemical physics}, 21(6):1087--1092, 1953.

\bibitem{neal2003}
Radford~M Neal.
\newblock Slice sampling.
\newblock {\em The Annals of Statistics}, 31(3):705--741, 2003.

\bibitem{graham2017}
Matthew~M. Graham and Amos~J. Storkey.
\newblock Asymptotically exact inference in differentiable generative models.
\newblock {\em Electronic Journal of Statistics}, 11(2):5105--5164, 2017.

\bibitem{le2016}
Tuan~Anh Le, Atilim~Gunes Baydin, and Frank Wood.
\newblock Inference compilation and universal probabilistic programming.
\newblock {\em Proceedings of the 20th International Conference on Artificial
  Intelligence and Statistics (AISTATS) 2017}, 54, 2017.

\bibitem{baydin2020}
Atilim~G{\"u}ne{\c{s}} Baydin, Lei Shao, Wahid Bhimji, Lukas Heinrich, Lawrence
  Meadows, Jialin Liu, Andreas Munk, Saeid Naderiparizi, Bradley Gram-Hansen,
  Gilles Louppe, et~al.
\newblock Etalumis: bringing probabilistic programming to scientific simulators
  at scale.
\newblock In {\em Proceedings of the International Conference for High
  Performance Computing, Networking, Storage and Analysis}, pages 1--24, 2019.

\bibitem{cranmer2019}
Kyle Cranmer, Johann Brehmer, and Gilles Louppe.
\newblock The frontier of simulation-based inference.
\newblock {\em Proceedings of the National Academy of Sciences}, 2020.

\bibitem{pritchard1999}
Jonathan~K Pritchard, Mark~T Seielstad, Anna Perez-Lezaun, and Marcus~W
  Feldman.
\newblock Population growth of human y chromosomes: a study of y chromosome
  microsatellites.
\newblock {\em Molecular biology and evolution}, 16(12):1791--1798, 1999.

\bibitem{sisson2007}
Scott~A Sisson, Yanan Fan, and Mark~M Tanaka.
\newblock Sequential monte carlo without likelihoods.
\newblock {\em Proceedings of the National Academy of Sciences},
  104(6):1760--1765, 2007.

\bibitem{papamakarios2016}
George Papamakarios and Iain Murray.
\newblock Fast $\epsilon$-free inference of simulation models with bayesian
  conditional density estimation.
\newblock In {\em Advances in Neural Information Processing Systems 29}, pages
  1028--1036. 2016.

\bibitem{lueckmann2017}
Jan-Matthis Lueckmann, Pedro~J Goncalves, Giacomo Bassetto, Kaan \"{O}cal,
  Marcel Nonnenmacher, and Jakob~H Macke.
\newblock Flexible statistical inference for mechanistic models of neural
  dynamics.
\newblock In {\em Advances in Neural Information Processing Systems 30}, pages
  1289--1299. 2017.

\bibitem{papamakarios2019a}
George Papamakarios, David Sterratt, and Iain Murray.
\newblock Sequential neural likelihood: Fast likelihood-free inference with
  autoregressive flows.
\newblock {\em Proceedings of Machine Learning Research}, 89:837--848, 2019.

\bibitem{greenberg2019}
David Greenberg, Marcel Nonnenmacher, and Jakob Macke.
\newblock Automatic posterior transformation for likelihood-free inference.
\newblock In {\em Proceedings of the 36th International Conference on Machine
  Learning}, volume~97 of {\em Proceedings of Machine Learning Research}, pages
  2404--2414. PMLR, 2019.

\bibitem{hermans2019}
Joeri Hermans, Volodimir Begy, and Gilles Louppe.
\newblock Likelihood-free mcmc with approximate likelihood ratios.
\newblock In {\em Proceedings of the 37th International Conference on Machine
  Learning}, volume~98 of {\em Proceedings of Machine Learning Research}. PMLR,
  2020.

\bibitem{elfi2018}
Jarno Lintusaari, Henri Vuollekoski, Antti Kangasr{\"a}{\"a}si{\"o}, Kusti
  Skyt{\'e}n, Marko J{\"a}rvenp{\"a}{\"a}, Pekka Marttinen, Michael~U Gutmann,
  Aki Vehtari, Jukka Corander, and Samuel Kaski.
\newblock Elfi: engine for likelihood-free inference.
\newblock {\em The Journal of Machine Learning Research}, 19(1):643--649, 2018.

\bibitem{gutmann2015}
Michael~U Gutmann and Jukka Corander.
\newblock Bayesian optimization for likelihood-free inference of
  simulator-based statistical models.
\newblock {\em The Journal of Machine Learning Research}, 17(1):4256--4302,
  2016.

\bibitem{louppe2016}
Gilles Louppe, Kyle Cranmer, and Juan Pavez.
\newblock carl: a likelihood-free inference toolbox.
\newblock {\em Journal of Open Source Software}, 1(1):11, 2016.

\bibitem{cranmer2015carl}
Kyle Cranmer, Juan Pavez, and Gilles Louppe.
\newblock Approximating likelihood ratios with calibrated discriminative
  classifiers.
\newblock {\em arXiv preprint arXiv:1506.02169}, 2015.

\bibitem{hypothesis-repo}
Joeri Hermans.
\newblock Hypothesis.
\newblock \url{https://github.com/montefiore-ai/hypothesis}, 2019.

\bibitem{pydelfi-repo}
Justin Alsing.
\newblock pydelfi: Density estimation likelihood-free inference.
\newblock \url{https://github.com/justinalsing/pydelfi}, 2019.

\bibitem{Klinger2018}
Emmanuel Klinger, Dennis Rickert, and Jan Hasenauer.
\newblock pyabc: distributed, likelihood-free inference.
\newblock {\em Bioinformatics}, 34(20):3591--3593, 2018.

\bibitem{abcpy-repo}
The ABCpy~authors.
\newblock Abcpy.
\newblock \url{https://github.com/eth-cscs/abcpy}, 2017.

\bibitem{paszke2019}
Adam Paszke, Sam Gross, Francisco Massa, Adam Lerer, James Bradbury, Gregory
  Chanan, Trevor Killeen, Zeming Lin, Natalia Gimelshein, Luca Antiga, Alban
  Desmaison, Andreas Kopf, Edward Yang, Zachary DeVito, Martin Raison, Alykhan
  Tejani, Sasank Chilamkurthy, Benoit Steiner, Lu~Fang, Junjie Bai, and Soumith
  Chintala.
\newblock Pytorch: An imperative style, high-performance deep learning library.
\newblock pages 8024--8035. 2019.

\bibitem{nflows-repo}
Conor Durkan, Artur Bekasov, George Papamakarios, and Iain Murray.
\newblock nflows: Normalizing flows in pytorch.
\newblock \url{https://github.com/bayesiains/nflows}, 2019.

\bibitem{delfi-repo}
mackelab.org.
\newblock Delfi: Density estimation likelihood-free inference.
\newblock \url{https://github.com/mackelab/delfi}, 2017.

\bibitem{theano}
Rami Al-Rfou, Guillaume Alain, Amjad Almahairi, Christof Angermueller, Dzmitry
  Bahdanau, Nicolas Ballas, Fr{\'e}d{\'e}ric Bastien, Justin Bayer, Anatoly
  Belikov, Alexander Belopolsky, et~al.
\newblock Theano: A python framework for fast computation of mathematical
  expressions.
\newblock {\em arXiv}, pages arXiv--1605, 2016.

\bibitem{goncalves2019}
Pedro~J Gon{\c{c}}alves, Jan-Matthis Lueckmann, Michael Deistler, Marcel
  Nonnenmacher, Kaan {\"O}cal, Giacomo Bassetto, Chaitanya Chintaluri,
  William~F Podlaski, Sara~A Haddad, Tim~P Vogels, David~S. Greenberg, and
  Jakob~H. Macke.
\newblock Training deep neural density estimators to identify mechanistic
  models of neural dynamics.
\newblock {\em bioRxiv}, page 838383, 2019.

\bibitem{lfi-repo}
Conor Durkan.
\newblock lfi.
\newblock \url{https://github.com/conormdurkan/lfi}, 2020.

\bibitem{durkan2020}
Conor Durkan, Iain Murray, and George Papamakarios.
\newblock On contrastive learning for likelihood-free inference.
\newblock {\em Proceedings of the 36th International Conference on Machine
  Learning}, 98, 2020.

\bibitem{joblib}
Gael Varoquaux.
\newblock joblib.
\newblock \url{https://github.com/joblib/joblib}, 2008.

\end{thebibliography}

\end{document}